\documentclass[runningheads]{llncs}

\usepackage{eccv}

\usepackage{eccvabbrv}

\usepackage{multirow}
\usepackage{graphicx}
\usepackage{booktabs}
\usepackage{wrapfig}
\usepackage{xcolor}
\usepackage{marvosym}
\usepackage{colortbl} 
\definecolor{highlight}{gray}{0.9} 
\usepackage[accsupp]{axessibility}  
\usepackage[most]{tcolorbox}

\usepackage[pagebackref,breaklinks,colorlinks,citecolor=eccvblue]{hyperref}

\usepackage{orcidlink}

\begin{document}

\title{SegPAR: Class-Centric Decision-Based Sparse Attack for Semantic Segmentation} 

\titlerunning{SegPAR: Class-Centric Sparse Attack}

\author{Dongsu Song\inst{1,3}\orcidlink{0009-0006-1443-5736} \and
DaeYun GO\inst{1,2}\orcidlink{0009-0000-2679-525X} \and
Boseung Seo\inst{1,2}\orcidlink{0009-0004-7915-7664} \and
Jay Hoon Jung\inst{2}\textsuperscript{\Letter}\orcidlink{0000-0002-9495-0693}}

\authorrunning{D. Song et al.}

\institute{Interdisciplinary Program in Space Systems Engineering, Korea Aerospace University, South Korea 
\and
Department of Artificial Intelligence, Korea Aerospace University, South Korea
\and
Department of Computer Science, Korea Aerospace University, South Korea
}

\maketitle
\begingroup
\renewcommand{\thefootnote}{}
\footnotetext{
\textsuperscript{\Letter}\ Corresponding \& first authors:
\href{mailto:jhjung@kau.ac.kr}{jhjung@kau.ac.kr} \& 
\href{mailto:raister01@kau.kr}{raister01@kau.kr}.}
\endgroup

\begin{abstract}
Despite the practical relevance of sparse decision-based black-box threats, they have received limited attention in semantic segmentation. To bridge this gap, we adapt the most representative decision-based black-box sparse attacks from the classification domain to serve as baselines, establishing a rigorous benchmark for this underexplored setting. In this context, we demonstrate that one of the existing methods suffers from severe query inefficiency due to its image-centric pixel accumulation, which rapidly exhausts query budgets across the vast image space. To overcome this, we propose SegPAR, a novel decision-based framework that shifts to a class-centric exploration paradigm. Furthermore, to eliminate the misleading feedback generated by standard decision rewards during pixel accumulation, we introduce a novel discrepancy reward. Extensive experiments show that SegPAR significantly outperforms black-box baselines in sparsity efficiency and MIoU reduction, while remaining competitive with white-box sparse attacks. Code is available at \href{https://github.com/KAU-QuantumAILab/SegPAR}{https://github.com/KAU-QuantumAILab/SegPAR}.
  \keywords{Black-box \and Sparse Attack  \and Class-centric}
\end{abstract}

\begin{figure}[tb]
  \centering
  \includegraphics[width=1\textwidth]{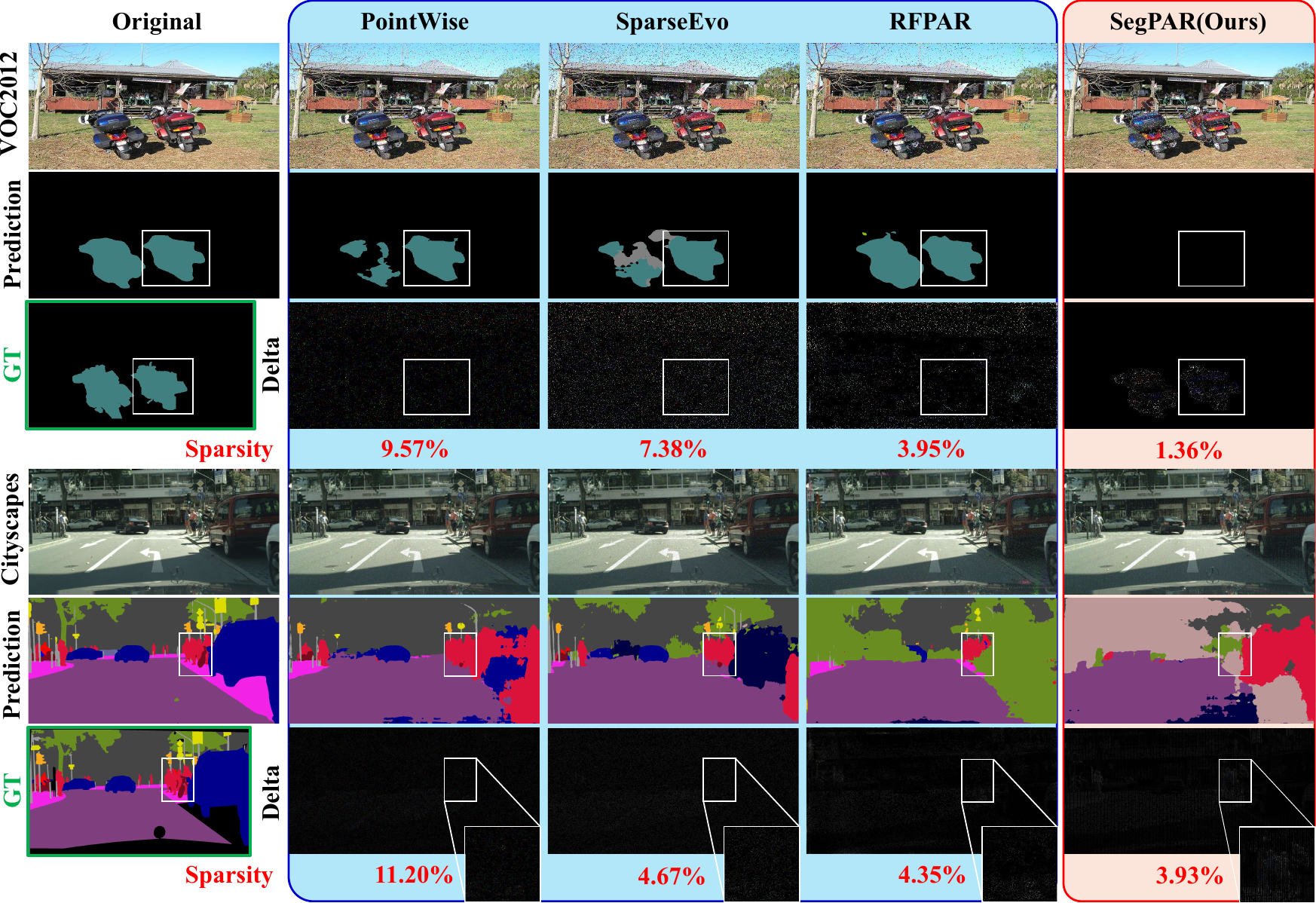}
  \caption{\textbf{Qualitative comparison of sparse adversarial attacks on semantic segmentation models.} Our proposed SegPAR induces severe misclassification with significantly lower sparsity compared to the baselines. As shown in the white boxes of the Delta maps, SegPAR maximizes attack efficiency by precisely targeting critical pixels, resulting in fatal semantic disruptions with minimal perturbations.}
  \label{fig:teaser}
\end{figure}

\section{Introduction}
\label{sec:intro}
Deep learning models have achieved remarkable success across domains \cite{chatgpt, gemini2.5, nvidia, SAM3, waymo}. However, prior work has shown that these models are susceptible to adversarial attacks, which introduce imperceptible perturbations that cause incorrect outputs \cite{att_ref1, att_ref2, att_ref3, att_ref4}. These attacks are generally categorized into \textit{white-box} and \textit{black-box} settings based on the adversary's knowledge \cite{white_black}. White-box attacks assume access to comprehensive model information, including gradients, architectures, and internal features \cite{pgd, white1}. Conversely, black-box attacks operate under limited information, relying solely on input-output interactions such as decision labels and confidence scores \cite{black1, black2, black3}. Given that black-box scenarios more accurately reflect the constraints of real-world deployments, advancing the study of black-box attacks is of critical importance.

Black-box attacks are typically categorized into \textit{transfer-based} and \textit{query-based} approaches. Transfer-based attacks craft adversarial perturbations on surrogate models and rely on cross-model transferability, whereas query-based attacks interact with the victim model through iterative perturbations~\cite{transfer1, query1}. Query-based attacks are particularly compelling because they are model-agnostic and do not require a representative surrogate. Within this paradigm, \textit{sparse} attacks that perturb only a small set of pixels have emerged as a challenging yet practically relevant setting~\cite{onepixel, pixle, sparse-rs}. Identifying influential pixels is NP-hard, and the combinatorial search space makes query-efficient optimization especially difficult under realistic query budgets~\cite{nphard1, nphard2}. Importantly, sparse threats can reflect real-world failure sources such as sensor defects (\eg, hot/dead pixels) in object detection pipelines and localized physical perturbations (\eg, stickers on road signs) that mislead autonomous driving systems~\cite{sparse1, sparse2, sparse3, rfpar}.

Despite the practical relevance of sparse decision-based black-box threats, they have received limited attention in semantic segmentation, where the search space scales with dense, multi-class outputs~\cite{croce2024towards}. 
While decision-based black-box attacks for semantic segmentation have recently begun to be explored~\cite{chen2024delving}, they largely focus on dense $\ell_p$-bounded perturbations and do not address explicit pixel-level sparsity. 
To bridge this gap, we adapt the most representative \textit{decision-based black-box sparse attacks} from the classification domain to serve as baselines, establishing a rigorous benchmark for this underexplored setting.
In this context, we demonstrate the existing method, RFPAR \cite{rfpar}, suffers from severe query inefficiency when applied to segmentation, as its image-centric pixel accumulation tends to concentrate queries on a few easy-to-exploit regions, failing to cover heterogeneous vulnerabilities across multiple semantic classes, thereby wasting the limited query budget. To overcome this, we propose Segmentation Pixel Attack using RL (\textbf{SegPAR}), a novel decision-based framework that shifts to a class-centric exploration paradigm. Furthermore, to eliminate the misleading feedback generated by standard decision rewards during pixel accumulation, we introduce a novel \textbf{discrepancy reward}. Extensive experiments show that SegPAR significantly outperforms black-box baselines in sparsity efficiency and MIoU reduction, as shown in \cref{fig:teaser}, while remaining competitive with white-box attacks.

In summary, our contributions are:
\begin{itemize}
    \item To the best of our knowledge, this work is the first systematic study of decision-based black-box \textbf{sparse} attacks for semantic segmentation.
    \item We propose \textbf{SegPAR}, a decision-based black-box sparse attack for semantic segmentation that improves query efficiency via class-centric exploration.
    \item We introduce a \textbf{discrepancy reward} that mitigates the unreliable signals of standard decision rewards during pixel accumulation. Combined with SegPAR, it achieves the strongest black-box performance and is competitive with white-box sparse baselines under the same sparsity constraints.
\end{itemize}

\section{Preliminaries} 
\label{sec:pre}
In our semantic segmentation attack setting, let $\boldsymbol{x}\in\{0,\dots,255\}^{C\times H \times W}$ be an input image and $f(\cdot)$ be a victim model outputting a predicted label map $\boldsymbol{y} \in \{0,\dots,K-1\}^{H\times W}$, where $H$, $W$, and $K$ denote the height, width, and number of classes, respectively. Our goal is to degrade the model's MIoU by perturbing only a small number of pixels under a limited query budget. Directly optimizing MIoU in a black-box setting is challenging since it is a non-smooth, set-based metric. Therefore, we propose an effective proxy: maximizing the fraction of misclassified pixels under a sparsity constraint. The problem is formulated as:
\begin{align}
  \max_{\boldsymbol{\delta}} \;\; \frac{1}{WH}\sum_{i=0}^{H-1}\sum_{j=0}^{W-1} \mathbb{I}\!\left[f(\boldsymbol{\hat{x}})_{i,j} \neq f(\boldsymbol{x})_{i,j}\right],\quad \text{s.t.}\quad \|\boldsymbol{\delta}\|_0 = \|\boldsymbol{\hat{x}}-\boldsymbol{x}\|_0 \le \epsilon,  
\end{align}
where $\boldsymbol{\hat{x}}$ is the modified image, $\|\cdot\|_0$ is defined as the number of modified pixels, and $\mathbb{I}$ is the indicator function. Under our decision-based black-box setting, the optimization relies strictly on label queries, while ground-truth annotations are reserved exclusively for final evaluation.

\section{Related Works}
\subsection{Decision-based Sparse Attack}
\label{sec:decision-based sparse attack}
Decision-based sparse attacks seek sparse perturbations while maximizing attack success using only the model's final hard-label decisions. Representative methods in this category include Pointwise \cite{pointwise} and SparseEvo \cite{SparseEvo}. Pointwise \cite{pointwise}, a decision-based sparse attack, minimizes perturbed pixels by initially inducing misclassification and using a greedy search to revert pixels. To reduce its search space complexity, SparseEvo \cite{SparseEvo} optimizes a binary vector of perturbed locations via an evolutionary algorithm. Since both require an initial successful attack, adapting them to semantic segmentation necessitates redefining ``attack success'' using the \textbf{Success Ratio (SR)}. The SR is defined as:
$
    \mathrm{SR}(x,\hat{x}) = \frac{1}{WH}\sum_{i=0}^{H-1}\sum_{j=0}^{W-1} \mathbb{I}\!\left[f(\hat{x})_{i,j} \neq f(x)_{i,j}\right].
$
Perturbations are removed only if the attack succeeds ($\mathrm{SR} \ge \tau$).
\begin{figure}[t]
  \centering
   \includegraphics[width=1\textwidth]{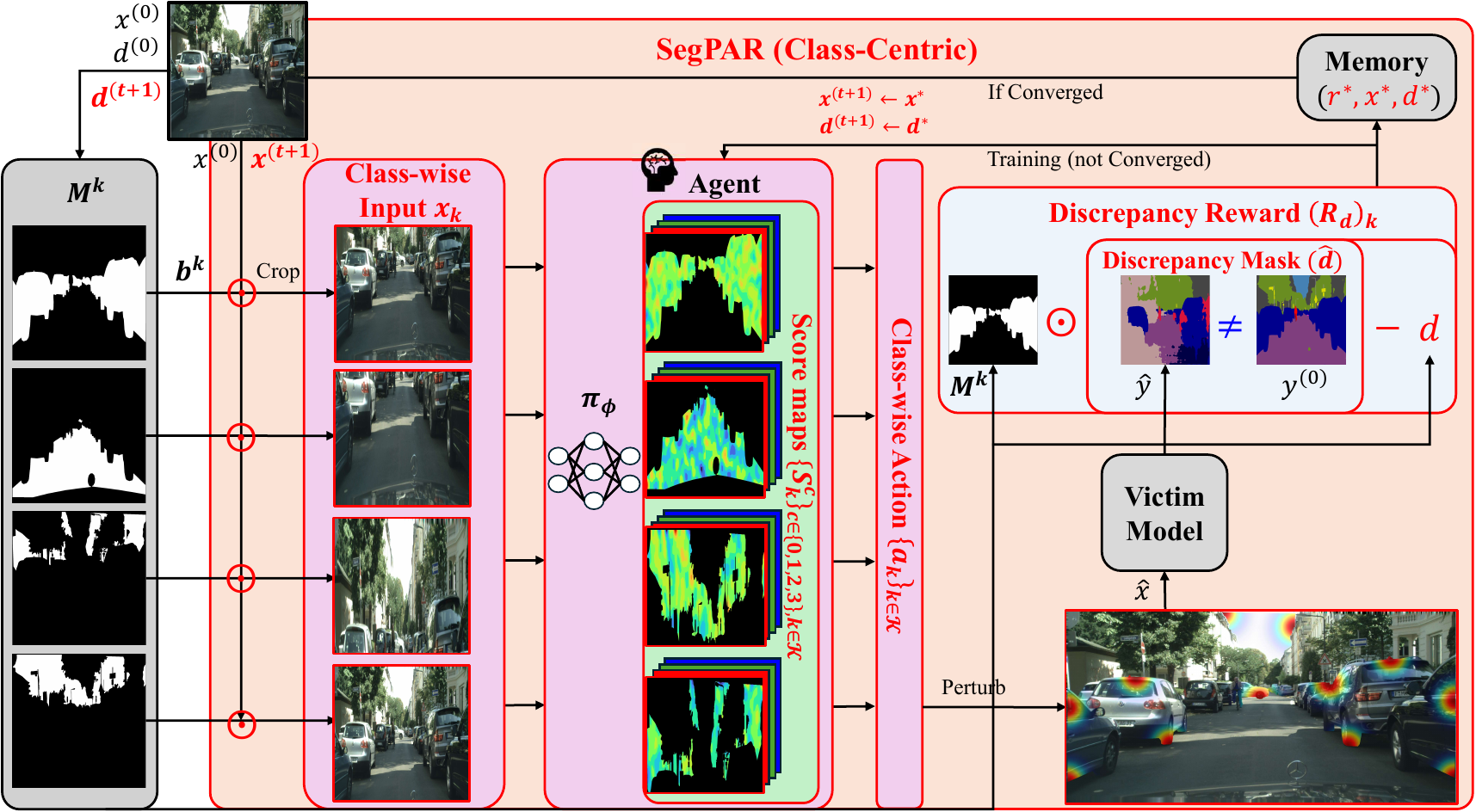}
\caption{\textbf{SegPAR Framework.}
At initialization, SegPAR extracts fixed per-class masks $\{M^k\}$ from the initial prediction $y^{(0)}$ and keeps them throughout the episode. 
At each step $t$, class-wise inputs are constructed by cropping the current image, $x_c^{(t)}=\mathrm{Crop}(x^{(t)}, b^k)$, where $b^k$ is the crop box for class $k$. 
Conditioned on the class and cropped input, the RL agent outputs sparse pixel perturbations to generate $\hat{x}^{(t)}$, which is then queried on the victim model $f$. 
The discrepancy mask is computed against the initial prediction as $d^{(t)}=\mathbf{1}[f(\hat{x}^{(t)}) \neq y^{(0)}]$. 
The class-wise discrepancy reward is defined within each fixed class mask, $R_d^k \propto \sum_{i,j} M_{i,j}^k \bigl(d_{i,j}^{(t)} - d_{i,j}^{(t-1)}\bigr)$. 
During training, SegPAR stores the best perturbation, discrepancy mask, and reward in memory; after convergence, it updates $x^{(t+1)} \leftarrow x^*$ and $d^{(t+1)} \leftarrow d^*$.
}
\label{fig:framework}
\end{figure}

\subsection{Score-based Sparse Attack}
Score-based sparse attacks seek sparse perturbations while maximizing attack success using only scores. While OnePixel \cite{onepixel} utilizes Differential Evolution for single-pixel manipulation, subsequent approaches adopt iterative accumulation strategies. Specifically, Sparse-RS \cite{sparse-rs} employs Random Search to progressively update and accumulate perturbations, and Pixle \cite{pixle} minimizes loss by iteratively rearranging neighboring pixel patches. Notably, RFPAR~\cite{rfpar} employs reinforcement learning. In RFPAR, an agent iteratively explores candidates $\hat{x}$, storing the best reward and attack parameters in memory. Upon convergence—when reward improvement stalls—the agent reinitializes and proceeds to the next step. We adapt this framework for a stricter decision-based black-box setting in semantic segmentation by eliminating the score component. Relying exclusively on decision rewards, we evaluate the candidate $\hat{x}$ using the \textbf{standard reward}: 
$R_{s}(x^{(t)},\hat{x}^{(t)}) = \frac{1}{o^2\cdot\|\hat{x}^{(t)}-x^{(t)}\|_0}\sum_{i=0}^{H-1}\sum_{j=0}^{W-1}\mathbb{I}\!\left[f(\hat{x}^{(t)})_{i,j} \neq f(x^{(t)})_{i,j}\right],$ 
 where $o=5$ is a scaling factor, and $x^{(t)}$ denotes the input image at the current step $t$.

\subsection{White-box Sparse Attack}
White-box sparse attacks leverage model gradients to achieve misclassification under a sparsity constraint. Early methods such as JSMA\cite{JSMA}, CW-$l_0$\cite{att_ref4}, and SparseFool\cite{nphard1} focused on inducing sparsity through direct optimization or surrogate $l_1$ norms. While PGD$_0$\cite{PGD0} extends Projected Gradient Descent to the $l_0$ space, it is prone to local optima due to the non-convex nature of the $l_0$ constraint. To mitigate this, sPGD\cite{sPGD} was introduced, which decomposes the perturbation $\delta$ into a magnitude tensor $p$ and a location mask. By employing continuous surrogate variables and top-$k$ projection, sPGD overcomes the non-differentiability of the $l_0$ norm. To further boost optimization, sPGD updates $p$ using both the \textit{projected} gradient restricted by the mask and the \textit{unprojected} gradient that ignores the mask.

\section{Method}
\label{sec:method}
In this section, we present the overall framework of SegPAR, illustrated in \cref{fig:framework}. SegPAR reformulates RFPAR~\cite{rfpar} into a class-wise attack paradigm tailored for semantic segmentation. We then analyze the limitations of the conventional pixel-accumulation reward and propose the discrepancy reward to address these issues and improve attack performance.
\subsection{SegPAR}
\label{sec:segpar}
\subsubsection{Insight.}
Deep neural networks partition the input space into complex decision regions, which in ReLU-based networks can be locally viewed as piece-wise linear boundary facets~\cite{jung2020boundaries}. Regardless of the exact boundary geometry, semantic segmentation requires predictions at every pixel. Thus, an attack must induce label flips by crossing a collection of heterogeneous, pixel- and class-specific boundaries, rather than a single global boundary.
Consequently, an image-centric exploration strategy can quickly concentrate queries on a few high-reward (often large or easily perturbed) regions, repeatedly probing similar areas while leaving other class-specific vulnerabilities under-explored.
We therefore adopt a class-centric exploration strategy that conditions the RL agent on each class present in the image, encouraging perturbations to be distributed across diverse regions and improving the efficiency of reaching relevant decision boundaries.
\subsubsection{Class-wise Input.}
\label{sec:class-wise}
To address this issue, we shift the original RFPAR paradigm, which generates an action from the entire image as the state, to a class-wise paradigm that uses a per-class state representation. 
Given an original image $\boldsymbol{x}$, we first query the victim model to obtain its prediction.
From $f(\boldsymbol{x})$, we construct and fix a set of class-wise binary masks $\mathcal{M}=\{M^{k}\}_{k\in\mathcal{K}}$, where $M^{k}\in\{0,1\}^{H\times W}$ indicates the predicted pixels belonging to class $k$ and $\mathcal{K}$ is defined as the set of classes that appear in the prediction for the given image:
\begin{equation}
    M^{k}_{i,j} = \mathbb{I}\!\left[f(\boldsymbol{x})_{i,j}=k\right], \quad \forall (i,j)\in\{0,\dots,H-1\}\times\{0,\dots,W-1\}.
\end{equation}
For each mask $M^{k}$, we define a bounding-box function $B(\cdot): M^k \rightarrow \mathbb{N}^{4},$
which returns the tightest axis-aligned bounding box that encloses the foreground pixels of the mask. Specifically, the bounding box for class $k$ is
$b^{k}=B(M^{k})=\big(i^{k}_{\min},\, j^{k}_{\min},\, i^{k}_{\max},\, j^{k}_{\max}\big)$,
where $i$ and $j$ denote the vertical and horizontal image coordinates, respectively.
Using $b^{k}$, we extract a class-specific crop from the original image and use it as the agent state $\boldsymbol{x}_{k} = \mathrm{Crop}\!\left(\boldsymbol{x},\, b^{k}\right).$
In summary, instead of using the full image $x$ as the state, the agent policy $\pi_\phi$ operates on a per-class state $x_k$ constructed from the model’s predicted region for class $k$.

\subsubsection{Class-wise Action.}
\label{sec:action}

As we convert the agent state into a class-specific region, the action space must be redesigned accordingly.
RFPAR samples coordinates from a Gaussian distribution, which implicitly assumes a rectangular target region.
However, class-wise masks are typically irregular; applying the same strategy can induce undesired behavior.
In particular, a bounding box often includes pixels from other classes, so perturbing and evaluating rewards over the entire box can contaminate the signal with non-target effects, destabilizing policy learning for the target class $k$.

To resolve this issue, we propose a mask-based sampling strategy: instead of sampling coordinates from a Gaussian, the agent predicts a probability map over candidate attack locations and samples from the induced distribution.
Specifically, given the state $\boldsymbol{x}_k$, the agent produces a 4-channel score map
$\boldsymbol{S}_k \in \mathbb{R}^{4 \times h_k \times w_k}$,
where $h_k = i^{k}_{\max}-i^{k}_{\min} + 1$ and $w_k = j^{k}_{\max}-j^{k}_{\min} + 1$.
Let $\boldsymbol{m}^k = \mathrm{Crop}(M^{k}, b^{k})$ be the class mask cropped to the same region as $\boldsymbol{x}_k$.
Using the first channel $\boldsymbol{S}_k^{0}$ as spatial logits, we define a masked softmax distribution over locations:
\begin{equation}
    \boldsymbol{P}_k
    = \mathrm{Softmax}\!\left(
    \boldsymbol{S}_k^0 + (1-\boldsymbol{m}^k)\cdot (-10^9)
    \right),
\end{equation}
where the softmax is taken over all $N=h_k w_k$ spatial indices. This assigns zero probability to masked-out pixels, confining the search space to the actual class region regardless of the bounding box size. Let $\boldsymbol{p}_k=\mathrm{vec}(\boldsymbol{P}_k)\in[0,1]^N$.
We sample $n$ locations without replacement as in our implementation.
Let $\mathcal{I}=\{I_1,\dots,I_n\}\sim \mathrm{MultinomialNR}(\boldsymbol{p}_k; n)$ denote weighted sampling of $n$ indices without replacement.
Each sampled index $I_l$ is converted to 2D coordinates by
\( j_l = (I_l \bmod w_k) + j^{k}_{\min}\), 
\( i_l = \lfloor I_l / w_k \rfloor + i^{k}_{\min} \).

For the RGB decision, we use the remaining three channels as logits.
Let $\boldsymbol{S}_k^{c}\in\mathbb{R}^{h_k\times w_k}$ be the $c$-th RGB logit map for $c\in\{1,2,3\}$, and denote its vectorization by $\boldsymbol{s}^{c}=\mathrm{vec}(\boldsymbol{S}_k^{c})$.
Conditioned on each sampled location $I_l$, we independently sample Bernoulli variables $z_{l,c} \sim \mathrm{Bernoulli}(\text{Sigmoid}(s^c_{I_l}))$ for each channel $c \in \{1, 2, 3\}$, where $\text{Sigmoid}(\cdot)$ is the sigmoid function.
Finally, the sampled action set for class $k$ is
$\boldsymbol{a}_k \triangleq \Big\{\big((i_l,j_l),\, z_{l,1},z_{l,2},z_{l,3}\big)\Big\}_{l=1}^{n}$.
To synthesize $\hat{\boldsymbol{x}}$, we apply these actions to $\boldsymbol{x}$ by mapping the binary decisions to the $\{0,255\}$ color space, consistent with the baseline: $\hat{x}_{i_l,j_l,c} = 255 \cdot z_{l,c}.$
Once $\hat{\boldsymbol{x}}$ is synthesized, we obtain the victim prediction $f(\hat{\boldsymbol{x}})$.
When training the agent with $R_s$ at step $t$, the class-specific reward $(R_s)_k$ for each class $k$ is defined as 
\begin{equation}
   (R_s(x^{(t)},\hat{x}^{(t)}))_k  = \frac{1}{o^2\cdot n}\sum_{i=0}^{H-1}\sum_{j=0}^{W-1}M_{i,j}^k\cdot\mathbb{I}\!\left[f(\hat{x}^{(t)})_{i,j} \neq f(x^{(t)})_{i,j}\right], 
\end{equation}
thereby isolating the feedback and avoiding contamination from non-target classes.
The agent is subsequently trained in a batched manner by optimizing the policy with respect to the rewards $\{(R_s(x^{(t)},\hat{x}^{(t)}))_{k}\}_{k\in\mathcal{K}}$.

\subsection{Discrepancy Reward}
\subsubsection{Standard Reward Inefficiency.}
\begin{wrapfigure}[12]{r}{0.48\textwidth} 
  \centering
  \includegraphics[width=0.48\textwidth]{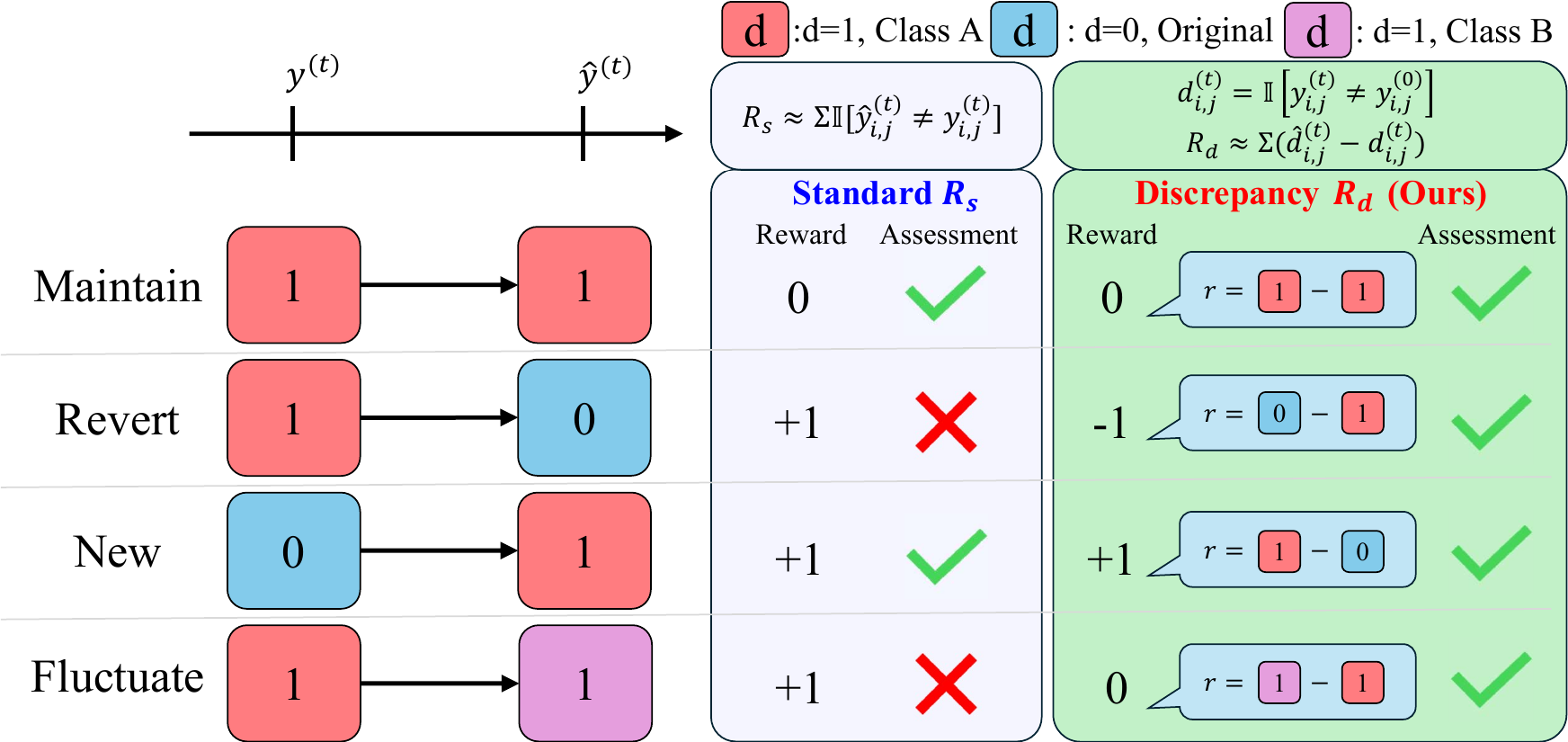}
  \caption{\textbf{Comparison of the standard reward and the discrepancy reward.}}
  \label{fig:discrepancy_reward}
\end{wrapfigure}
We analyze the inherent limitations of the standard reward function in the context of cumulative attack frameworks. Specifically, we investigate the inefficiencies that arise in tasks involving multiple target instances, such as object detection and semantic segmentation, where the agent must handle a significantly larger set of target pixels. Let $y_{i,j}^{(t)} = f(x^{(t)})_{i,j}$ be the prediction of the model for a pixel at spatial coordinate $(i, j)$ at step $t$, and let $y_{i,j}^{(0)} = f(x^{(0)})_{i,j}$ denote the original prediction class of that pixel. In cumulative attack frameworks, relying only on the prediction transition ($y_{i,j}^{(t)} \neq \hat{y}_{i,j}^{(t)}$, where $\hat{y}_{i,j}^{(t)} = f(\hat{x}^{(t)})_{i,j}$) severely misaligns the agent's optimization direction. To mathematically demonstrate this, we decompose the pixel-wise state transitions at step $t$ into four mutually exclusive subsets, leading to the $t+1$ state:

\begin{itemize}
    \item \textbf{Case 1 (Maintained Misclassification):} Pixels where a successful untargeted attack is stably maintained.
    \begin{equation}
        \mathcal{S}_{maintain}^{(t)} = \left\{ (i,j) \mid y_{i,j}^{(t)} \neq y_{i,j}^{(0)} \land \hat{y}_{i,j}^{(t)} = y_{i,j}^{(t)} \right\}
    \end{equation}
    \item \textbf{Case 2 (Reversion Failure):} Pixels where the accumulated noise interferes with a previously successful attack, reverting the prediction back to the original class.
    \begin{equation}
        \mathcal{S}_{revert}^{(t)} = \left\{ (i,j) \mid y_{i,j}^{(t)} \neq y_{i,j}^{(0)} \land \hat{y}_{i,j}^{(t)} = y_{i,j}^{(0)} \right\}
    \end{equation}
    \item \textbf{Case 3 (New Misclassification):} Pixels that are newly and successfully deviated from the original class.
    \begin{equation}
        \mathcal{S}_{new}^{(t)} = \left\{ (i,j) \mid y_{i,j}^{(t)} = y_{i,j}^{(0)} \land \hat{y}_{i,j}^{(t)} \neq y_{i,j}^{(0)} \right\}
    \end{equation}
    \item \textbf{Case 4 (Unstable Fluctuation):} Pixels that remain misclassified but fluctuate to a different incorrect class, indicating wasted perturbations.
    \begin{equation}
        \mathcal{S}_{fluctuate}^{(t)} = \left\{ (i,j) \mid y_{i,j}^{(t)} \neq y_{i,j}^{(0)} \land \hat{y}_{i,j}^{(t)} \neq y_{i,j}^{(0)} \land y_{i,j}^{(t)} \neq \hat{y}_{i,j}^{(t)} \right\}
    \end{equation}
\end{itemize}

Applying $R_{s}$ to these subsets reveals critical optimization flaws. 
Since $R_{s}$ is a transition-based reward, it functionally depends on the number of pixels whose predictions change between steps, \ie,
\begin{equation}
R_s(x^{(t)},\hat{x}^{(t)}) \propto\sum_{i,j}\mathbb{I}\!\left[y_{i,j}^{(t)} \neq \hat{y}_{i,j}^{(t)}\right]
= \left|\mathcal{S}_{revert}^{(t)}\right| + \left|\mathcal{S}_{new}^{(t)}\right| + \left|\mathcal{S}_{fluctuate}^{(t)}\right|.
\end{equation}
Therefore, the agent receives positive reinforcement not only for desirable transitions in $\mathcal{S}_{new}^{(t)}$ but also for undesirable ones in $\mathcal{S}_{revert}^{(t)}$ and $\mathcal{S}_{fluctuate}^{(t)}$. 
In particular, when a previously misclassified pixel reverts to the original class ($\mathcal{S}_{revert}^{(t)}$) or oscillates among incorrect classes ($\mathcal{S}_{fluctuate}^{(t)}$), $R_{s}$ still assigns a positive reward despite reducing the cumulative attack objective. 
This misleading feedback biases the agent toward suboptimal regions, increasing query complexity and causing redundant pixel perturbations.

\subsubsection{Discrepancy Mask-based Reward Function.}
\label{subsec:proposed_reward}

To overcome the limitations of the naive approach, we introduce a \textit{Discrepancy Mask}, denoted as $d^{(t)}$, to evaluate the attack success state at step $t$ relative to the initial prediction. For each pixel $(i,j)$, $d_{i,j}^{(t)}$ is defined as a binary indicator of attack success:
$d_{i,j}^{(t)} = \mathbb{I}(y_{i,j}^{(t)} \neq y_{i,j}^{(0)}).$
Based on this mask, we formulate our proposed reward function, $(R_{d}(x^{(t)},\hat{x}^{(t)}))_k$, as between the current discrepancy mask and the candidate discrepancy mask at step $t$:
\begin{equation}
    (R_{d}(x^{(t)},\hat{x}^{(t)}))_k =\frac{1}{o^{2}\cdot n} \sum_{i,j} M_{i,j}^k\cdot( \hat{d}_{i,j}^{(t)} - d_{i,j}^{(t)} ).
\end{equation}

This concise formulation resolves the key contradictions of $R_{s}$ across all four subsets. With the pixel-wise reward $r=\hat{d}_{i,j}^{(t)}-d_{i,j}^{(t)}$, we have $r=0$ for $\mathcal{S}_{maintain}$, $r=-1$ for $\mathcal{S}_{revert}$, $r=+1$ for $\mathcal{S}_{new}$, and $r=0$ for $\mathcal{S}_{fluctuate}$, as illustrated in \cref{fig:discrepancy_reward}. Consequently, $R_d$ directly tracks the marginal gain of cumulatively successful pixels, encouraging the agent to monotonically expand misclassified regions without unnecessary perturbations.

\subsubsection{Training \& Termination.}
Following RFPAR~\cite{rfpar}, we replace $y^t$ with $d^t$ in memory and track the maximum mean reward $r^* = \frac{1}{|\mathcal{K}|} \sum_{k \in \mathcal{K}} (R_d)_k$, maintaining the best $(d^*, x^*)$. We adopt the same hyperparameter suite as RFPAR and cap each run at 100 steps. During each step, the agent updates its policy via REINFORCE~\cite{williams1992simple} using label-only queries; once the memory converges, the agent is reinitialized and starts the next step with a new input.

\section{Experiments}
\label{sec:experiments}
We evaluate SegPAR against multiple baselines under strict query and sparsity budgets. Our approach consistently outperforms all black-box methods, achieving larger MIoU drops with fewer perturbed pixels, and remains highly effective against adversarially trained models. Ablations on per-class degradation and pixel accumulation confirm that our class-centric design and discrepancy reward drive these improvements. Furthermore, SegPAR performs competitively with strong white-box sparse attacks, highlighting its practical efficacy under realistic decision-only constraints.

\begin{table}[t]
\normalsize
\centering
\caption{\textbf{Evaluation of sparse attack performance across semantic segmentation benchmarks.} (-) indicates that the experiments are omitted due to the lack of official pre-trained weights for VOC2012.}
\label{tab:model_performance}
\resizebox{\textwidth}{!}{
\begin{tabular}{ll|cccc|cccc|cccc}
\toprule
\multirow{2}{*}{\textbf{Model}} & \multirow{2}{*}{\textbf{Attack}} & \multicolumn{4}{c|}{\textbf{Cityscapes}} & \multicolumn{4}{c|}{\textbf{ADE20K}} & \multicolumn{4}{c}{\textbf{VOC2012}} \\
\cmidrule(lr){3-6} \cmidrule(lr){7-10} \cmidrule(lr){11-14}
 & & \textbf{MIoU} & \textbf{R.MIoU} & \textbf{Sparsity} & \textbf{Query} & \textbf{MIoU} & \textbf{R.MIoU} & \textbf{Sparsity} & \textbf{Query} & \textbf{MIoU} & \textbf{R.MIoU} & \textbf{Sparsity} & \textbf{Query} \\
\midrule
\multirow{4}{*}{DeepLabV3} 
 & PointWise & \multirow{4}{*}{0.798} & 0.433 & 5.72\% & 997.3 & \multirow{4}{*}{0.377} & 0.172 & 6.73\% & 993.5 & \multirow{4}{*}{0.861} & 0.669 & 5.91\% & 952.5 \\
 & SparseEvo & & 0.228 & 5.43\% & 988.8 & & 0.118 & 5.15\% & 997.3 & & 0.469 & 5.15\% & 998.8 \\
 & RFPAR$_{R_s}$     & & 0.176 & 4.51\%& 965.3 & &  0.106 & 4.50\% & 998.1 & & 0.466 & 3.92\%& 986.3 \\
 \rowcolor{highlight}&SegPAR$_{R_d}$     & & \textbf{0.101} & \textbf{3.33\%}& 993.3 & &  \textbf{0.080} & \textbf{3.29\%}& 966.2 & & \textbf{0.238} & \textbf{2.24\%}& 680.4 \\
\midrule
\multirow{4}{*}{PSPNet}     
 & PointWise & \multirow{4}{*}{0.793} & 0.391 & 5.78\% & 996.3 & \multirow{4}{*}{0.380} & 0.199 & 6.32\% & 983.6 & \multirow{4}{*}{0.860} & 0.704 & 4.85\% & 929.6 \\
 & SparseEvo & & 0.246 & 4.50\% & 987.0 & & 0.139 & 5.54\% & 997.0 & & 0.488 & 4.66\% & 998.7 \\
 & RFPAR$_{R_s}$     & & 0.153 & 4.61\%& 990.5 & & 0.143 & 4.36\% & 999.0 & & 0.529 & 3.99\%& 984.3 \\
 \rowcolor{highlight}& SegPAR$_{R_d}$    & & \textbf{0.057} & \textbf{3.53\%}& 919.0 & & \textbf{0.087} & \textbf{3.11\%} & 965.3 & & \textbf{0.240} & \textbf{2.29\%}& 679.1 \\
\midrule
\multirow{4}{*}{SegFormer}  
 & PointWise & \multirow{4}{*}{0.802} & 0.594 & 10.51\%& 997.9 & \multirow{4}{*}{0.412} & 0.316 & 10.1\%& 973.2 & \multirow{4}{*}{-}     & -     & -     & -      \\
 & SparseEvo & & 0.602 & 5.30\%& 994.3 & & 0.293 & 7.03\% & 996.0 & & -     & -     & -      \\
 & RFPAR$_{R_s}$     & & 0.566 & 4.51\%& 958.5 & & 0.300 & 4.29\% & 912.0 & & - & - & - \\
 \rowcolor{highlight}& SegPAR$_{R_d}$    & & \textbf{0.341} & \textbf{3.96\%}& 960.4 & & \textbf{0.209} &\textbf{3.60\%} & 987.0 & &  -    & -    & -     \\
\midrule
\multirow{4}{*}{SETR}       
 & PointWise & \multirow{4}{*}{0.780} & 0.607 & 9.14\% & 997.1 & \multirow{4}{*}{0.397} & 0.284 & 8.31\% & 966.5 & \multirow{4}{*}{-}     & -     & -     & -      \\
 & SparseEvo & & 0.602 & 5.71\% & 997.3 & & 0.270 & 5.20\% & 997.3 & & -     & -     & -      \\
 & RFPAR$_{R_s}$     & & 0.527 & 4.53\%& 962.9 & & 0.285 & 3.90\% & 996.5 & &  -    &   -   &  -    \\
\rowcolor{highlight} &SegPAR$_{R_d}$    & & \textbf{0.481} & \textbf{3.89\%}& 940.4 & &\textbf{0.249}  & \textbf{3.50\%}  & 990.9  & &  -   &  -   &  -    \\
\bottomrule
\end{tabular}}
\end{table}
\subsection{Evaluation of Decision-based Sparse Attacks}

\subsubsection{Attacks.}
We compare SegPAR$_{R_d}$ with PointWise\cite{pointwise}, SparseEvo\cite{SparseEvo}, and RFPAR~\cite{rfpar} under a strict budget of $1{,}000$ queries. For the RL-based methods (SegPAR and RFPAR), we use patience $1$--$2$ with convergence thresholds in $[10^{-2}, 3 \times 10^{-2}]$. To align with the $5\%$ sparsity target, RFPAR enforces a global $5\%$ constraint over the entire image, whereas our class-centric SegPAR targets 5\% of pixels within each predicted class region (adjusted accordingly when the small number of classes makes it difficult to meet the overall 5\% sparsity). Since PointWise and SparseEvo are SR-parameterized, we sweep their SR values and report the run closest to the $5\%$ target. Note that these SR-based baselines may slightly exceed the $5\%$ constraint, which inherently favors them. Additional details are provided in the supplementary material.
\begin{table}[t]
\centering
\small
\caption{\textbf{Evaluation of sparse attack performance across adversarially trained semantic segmentation benchmarks.}}
\label{tab:defended_performance}
\resizebox{\textwidth}{!}{
\begin{tabular}{ll|cccc|cccc}
\toprule
\multirow{2}{*}{\textbf{Model}} & \multirow{2}{*}{\textbf{Attack}} & \multicolumn{4}{c|}{\textbf{Cityscapes}} & \multicolumn{4}{c}{\textbf{VOC2012}} \\
\cmidrule(lr){3-6} \cmidrule(lr){7-10}
 & & \textbf{MIoU} & \textbf{R.MIoU} & \textbf{Sparsity} & \textbf{Query} & \textbf{MIoU} & \textbf{R.MIoU} & \textbf{Sparsity} & \textbf{Query} \\
\midrule
\multirow{4}{*}{DeepLabV3\_DDCAT} 
 & PointWise & \multirow{4}{*}{0.710} & 0.502 & 25.56\% & 984.2  & \multirow{4}{*}{0.828} & 0.692 & 7.81\%  & 977.1 \\
 & SparseEvo &                        & 0.502 & 9.56\% & 910.2 &                        & 0.650 & 4.79\%  & 990.3 \\
 & RFPAR$_{R_s}$     &                        & 0.483 & 4.25\% & 840.0 &                        & 0.543 & 4.39\%  & 913.8  \\
 \rowcolor{highlight}& SegPAR$_{R_d}$    &                        & \textbf{0.428} & \textbf{4.24\%} & 878.3 &                        & \textbf{0.297} & \textbf{2.73\%} & 822.5 \\
\midrule
\multirow{4}{*}{DeepLabV3\_SAT}   
 & PointWise & \multirow{4}{*}{0.693}     & 0.495 & 23.17\%   & 993.8  & \multirow{4}{*}{0.775}     & 0.676 & 10.0\% & 918.6 \\
 & SparseEvo &                        & 0.468 & 8.42\%  & 932.9  &                        & 0.630 & 5.81\%  & 994.3 \\
 & RFPAR$_{R_s}$     &                        & 0.434 & 4.42\% & 851.4 &                        & 0.566 & 4.47\% & 948.5 \\
 \rowcolor{highlight}& SegPAR$_{R_d}$    &                        & \textbf{0.394} & \textbf{4.25\%} & 946.2 &                        & \textbf{0.326} & \textbf{2.92\%} & 799.3 \\
\midrule
\multirow{4}{*}{PSPNet\_DDCAT}    
 & PointWise & \multirow{4}{*}{0.714} & 0.523  & 23.06\% & 994.1  & \multirow{4}{*}{0.835} & 0.697 & 8.47\%  & 951.6 \\
 & SparseEvo &                        & 0.507 & 8.26\% & 939.0 &                        & 0.658 & 4.79\%  & 998.0 \\
 & RFPAR$_{R_s}$     &                        & 0.472 & 4.33\% & 873.0 &                        & 0.578 & 4.18\%  & 996.1 \\
 \rowcolor{highlight}& SegPAR$_{R_d}$    &                        & \textbf{0.405} & \textbf{4.22\%} & 935.3 &                        & \textbf{0.382} & \textbf{2.55\%} & 809.5 \\
\midrule
\multirow{4}{*}{PSPNet\_SAT}      
 & PointWise & \multirow{4}{*}{0.677}     & 0.495  & 24.34\%  & 986.7  & \multirow{4}{*}{0.891}     & 0.693 & 8.63\%  & 947.0 \\
 & SparseEvo &                        & 0.482  & 8.77\% & 922.7   &                        & 0.650 & 5.03\%  & 995.9 \\
 & RFPAR$_{R_s}$     &                        & 0.450 & \textbf{4.30\%} & 995.0 &                        & 0.591 & 4.43\% & 846.9 \\
 \rowcolor{highlight}& SegPAR$_{R_d}$    &                        & \textbf{0.384} & 4.47\% & 955.9 &                        & \textbf{0.332} & \textbf{2.79\%} & 832.6 \\
\bottomrule
\end{tabular}
}
\end{table}

\subsubsection{Victim models.}
To ensure the reproducibility of our experiments, we utilize pre-trained checkpoints provided by MMSegmentation. We select widely used models in the field of semantic segmentation, including DeepLabV3 \cite{deeplabv3}, PSPNet \cite{pspnet}, SegFormer \cite{segformer}, and SETR \cite{setr}. Specifically, DeepLabV3 and PSPNet are employed as CNN-based models, while SegFormer and SETR are utilized as Transformer-based architectures.

\subsubsection{Datasets.}
Due to the high computational cost of query-based attacks on dense prediction models, we evaluate on a fixed subset of 100 randomly sampled validation images from ADE20K\cite{ADE20K}, Pascal VOC2012\cite{pascal}, and Cityscapes\cite{cityscapes}. 
ADE20K contains 150 diverse classes with fine-grained part annotations, VOC2012 covers 20 common object categories, and Cityscapes focuses on urban driving scenes with 19 classes.

\subsubsection{Evaluation Metrics.}
We report Robust MIoU (R.MIoU), Sparsity, and Query to measure attack effectiveness and efficiency. R.MIoU is the victim model's MIoU on adversarial examples (lower is better). Sparsity is the average fraction of perturbed pixels (lower indicates fewer, more imperceptible changes). Query is the average number of forward passes required to obtain the final adversarial image (lower is more efficient).

\begin{figure}[tb]
  \centering
  \includegraphics[width=1\textwidth]{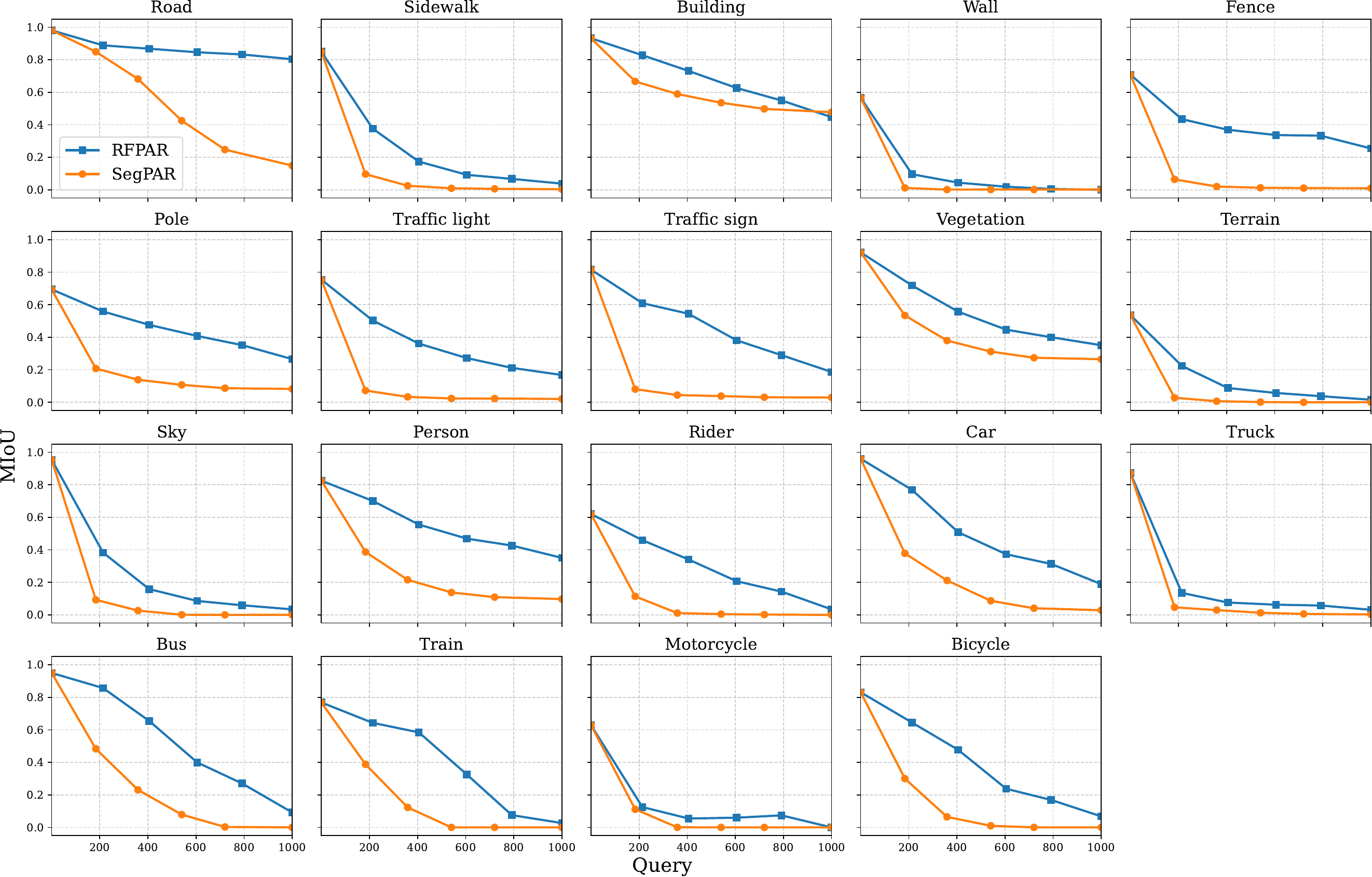}
  \caption{\textbf{Class-wise attack efficiency.} Comparison of MIoU degradation across individual classes in DeepLabV3. SegPAR demonstrates significantly faster and more effective attacks than RFPAR across all categories.}
  \label{fig:cityscapes}
\end{figure}
\subsubsection{Main Results.}
As shown in \cref{tab:model_performance}, SegPAR with $R_d$ consistently outperforms existing sparse attack techniques across all evaluated datasets, achieving the lowest R.MIoU under extreme sparsity at comparable or lower queries. For CNN-based architectures, SegPAR with $R_d$ dramatically reduces MIoU; notably, it drops the MIoU of PSPNet on Cityscapes to $\mathbf{0.057}$ and achieves extreme sparsity ($\mathbf{2.24\%}$) with significantly lower average queries (\eg, 680.42 for DeepLabV3) on VOC2012. Furthermore, our method proves highly effective even against Transformer-based models like SegFormer and SETR, which typically exhibit higher robustness against local perturbations due to their ability to capture global context. Despite this inherent robustness, our method maintains potent attack performance on Cityscapes using a sparsity of only $\mathbf{3.96\%}$ against SegFormer---a stark contrast to the $\mathbf{10.51\%}$ required by PointWise. Consistently observed across all benchmarks including ADE20K, these results confirm that our approach effectively accelerates the search process to precisely navigate the most vulnerable pixels, regardless of the target architecture.

\subsection{Evaluation on Adversarially Trained Models}
\cref{tab:defended_performance} reports attack performance against defended models (DDCAT\cite{ddcat}, SAT\cite{sat}). SegPAR consistently attains the lowest R.MIoU with lower sparsity than prior methods. Notably, adversarial training effectiveness varies across datasets, which we conjecture is partially influenced by image resolution. On high-resolution Cityscapes, the enlarged search space inherently hinders sparse attacks, allowing adversarial training to substantially boost robustness---even enabling CNNs to approach Transformer-level resilience. Conversely, on lower-resolution VOC2012, sparse attacks remain highly effective, yielding only marginal defense gains. While domain complexity also plays a role, current adversarial training appears to offer meaningful protection primarily in high-resolution settings, leaving lower-resolution regimes vulnerable.
\begin{table}[t]
\normalsize
\centering
\caption{Evaluation of Pixel-accumulating Adversarial Attacks across Different Losses.}
\label{tab:comparison with Loss}
\resizebox{\textwidth}{!}{
\begin{tabular}{lll|cccc|cccc|cccc}
\toprule
\multirow{2}{*}{\textbf{Model}} & \multirow{2}{*}{\textbf{Objective}} & \multirow{2}{*}{\textbf{Attack}} & \multicolumn{4}{c|}{\textbf{Cityscapes}} & \multicolumn{4}{c|}{\textbf{ADE20K}} & \multicolumn{4}{c}{\textbf{VOC2012}} \\
\cmidrule{4-15}
& & & \textbf{MIoU} & \textbf{R.MIoU} & \textbf{Sparsity} & \textbf{Query} & \textbf{MIoU} & \textbf{R.MIoU} & \textbf{Sparsity} & \textbf{Query} & \textbf{MIoU} & \textbf{R.MIoU} & \textbf{Sparsity} & \textbf{Query} \\
\midrule

\multirow{6}{*}{DeepLabV3} & \multirow{3}{*}{Standard} & Pixle & \multirow{3}{*}{0.798} & \textbf{0.394} & 4.69\% & 995.4 & \multirow{3}{*}{0.377} & 0.140 & 4.79\% & 995.5 & \multirow{3}{*}{0.861} & 0.447 & 4.77\% & 995.5 \\
& & Sparse-RS & & 0.172 & \textbf{4.87\%} & 998.7 & & 0.100 & \textbf{4.83\%} & 998.1 & & 0.423 & \textbf{4.84\%} & 998.2 \\
& & RFPAR & & \textbf{0.176} & 4.51\% & 993.3 & & 0.106 & 4.50\% & 998.1 & & 0.466 & 3.92\% & 986.3 \\
\cmidrule{2-15}
\rowcolor{highlight}&& Pixle &  & 0.411 & \textbf{4.04\%} & 995.3 & & \textbf{0.111} & \textbf{4.72\%} & 995.3 & & \textbf{0.279} & \textbf{4.59\%} & 994.9 \\
\rowcolor{highlight}& & Sparse-RS & & \textbf{0.152} & \textbf{4.87\%} & 998.5 & & \textbf{0.092} & \textbf{4.83\%} & 998.3 & & \textbf{0.258} & \textbf{4.84\%} & 998.0 \\
\rowcolor{highlight}&\multirow{-3}{*}{Discrepancy} & RFPAR &\multirow{-3}{*}{0.798} & 0.210 & \textbf{3.56\%} & 994.9 & \multirow{-3}{*}{0.377} & \textbf{0.090} & \textbf{3.48\%} & 947.8 & \multirow{-3}{*}{0.861} & \textbf{0.277} & \textbf{3.43\%} & 950.1 \\
\midrule

\multirow{6}{*}{PSPNet} & \multirow{3}{*}{Standard} & Pixle & \multirow{3}{*}{0.793} & \textbf{0.369} & 4.69\% & 995.8 & \multirow{3}{*}{0.380} & 0.168 & 4.79\% & 995.4 & \multirow{3}{*}{0.860} & 0.374 & 4.77\% & 995.2 \\
& & Sparse-RS & & 0.153 & \textbf{4.87\%} & 998.7 & & 0.121 & \textbf{4.83\%} & 998.3 & & 0.536 & \textbf{4.84\%} & 997.7 \\
& & RFPAR & & \textbf{0.153} & 4.61\% & 990.6 & & 0.143 & 4.36\% & 999.0 & & 0.529 & 4.00\% & 984.3 \\
\cmidrule{2-15}
\rowcolor{highlight}&  & Pixle &  & 0.383 & \textbf{4.13\%} & 995.8 & & \textbf{0.131} & \textbf{4.69\%} & 995.4 & \multirow{3}{*}{0.860} & \textbf{0.246} & \textbf{4.59\%} & 995.3 \\
\rowcolor{highlight}& & Sparse-RS & & \textbf{0.139} & \textbf{4.87\%} & 998.6 & & \textbf{0.092} & \textbf{4.83\%} & 998.1 & & \textbf{0.304} & \textbf{4.84\%} & 997.9 \\
\rowcolor{highlight}&\multirow{-3}{*}{Discrepancy} & RFPAR &\multirow{-3}{*}{0.793} & 0.171 & \textbf{3.92\%} & 993.0 &\multirow{-3}{*}{0.380} & \textbf{0.133} & \textbf{3.19\%} & 982.5 &\multirow{-3}{*}{0.860} & \textbf{0.295} & \textbf{3.13\%} & 936.7 \\
\bottomrule
\end{tabular}}
\end{table}

\subsection{Per-class MIoU Degradation Analysis}
\label{sec:per-class MIou degradation}
In this section, we evaluate the impact of class-centric exploration by comparing RFPAR and SegPAR, both employing $R_s$. As illustrated in \cref{fig:cityscapes}, SegPAR, which represents a paradigm shift from an image-centric to a class-centric approach, exhibits a significantly steeper degradation curve for most categories compared to RFPAR. These results indicate that our class-centric formulation, which constructs per-class inputs from the predicted regions and attacks multiple class-specific areas in parallel, is more effective than image-centric exploration that tends to concentrate queries on a few dominant regions. Notably, safety-critical classes for autonomous driving---such as traffic signs, traffic lights, persons, and riders---are compromised much earlier than with RFPAR, with SegPAR driving their per-class MIoU down sharply within $\sim$200 queries. This underscores that our method is not only numerically superior in attack performance, but also sufficiently powerful to cause practically meaningful degradation in real-world perception pipelines, even under strict black-box constraints.

\begin{figure}[t]
  \centering
  \includegraphics[width=1\textwidth]{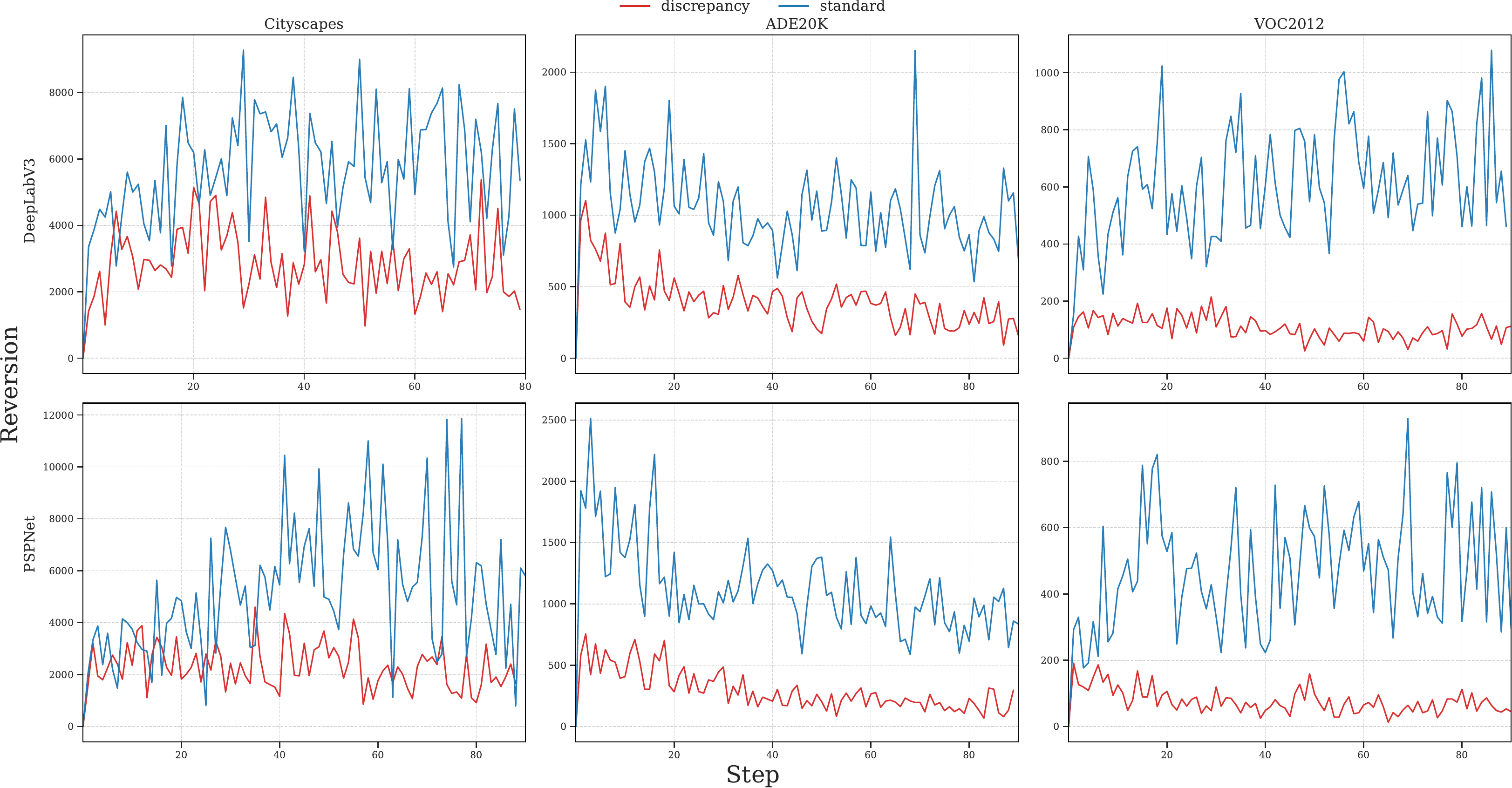}
  \caption{\textbf{Reduction of Reversion.} Our discrepancy reward effectively suppresses reversion, resolving a major hurdle in agent training caused by the standard reward.}
  \label{fig:reversion}
\end{figure}

\subsection{Evaluation of Discrepancy Reward}
While \cref{sec:per-class MIou degradation} already demonstrates the superiority of our class-centric architecture over the image-centric RFPAR under the same standard reward, this section isolates and explicitly evaluates the contribution of the proposed discrepancy reward $R_d$ by applying it to Pixle\cite{pixle} and Sparse-RS\cite{sparse-rs}. We optimize the discrepancy objective $\mathcal{L}_d=-R_d$ (equivalently maximizing $R_d$) and compare it with the standard objective $\mathcal{L}_s=-R_s$.
As shown in \cref{tab:comparison with Loss}, on ADE20K and VOC2012, the discrepancy objective leads to a larger reduction in MIoU than the standard objective under the same (or fewer) sparsity, indicating that $R_d$ enables more query-efficient pixel accumulation.
In contrast, on the high-resolution Cityscapes benchmark, applying $R_d$ to RFPAR can achieve lower sparsity but yields a smaller MIoU drop than the standard objective. We attribute this to the enlarged effective action space in high-resolution images: since $R_d$ provides non-zero reward only when the set of cumulatively successful pixels expands, informative reward events become rare, causing the search to stagnate before discovering pixels that maximize $R_d$.
As verified in \cref{tab:model_performance}, this limitation is largely alleviated when combining $R_d$ with our proposed SegPAR, which reduces the effective search space via class-wise inputs and mask-constrained exploration. Consequently, their combination achieves the largest MIoU reduction in high-resolution settings.
Furthermore, \cref{fig:reversion} shows that the discrepancy objective substantially suppresses reversion compared to the standard objective. We quantify reversion at step $t$ as $\left|\mathcal{S}_{revert}^{(t)}\right|$, \ie, the number of misclassified pixels that revert to the original prediction between steps. Overall, these results indicate that $R_d$ mitigates the reversion issue by discouraging counter-productive transitions and improving search efficiency.

\subsection{Comparison with White-box Sparse Baseline}
\begin{figure}[tb]
  \centering
  \includegraphics[width=1\textwidth]{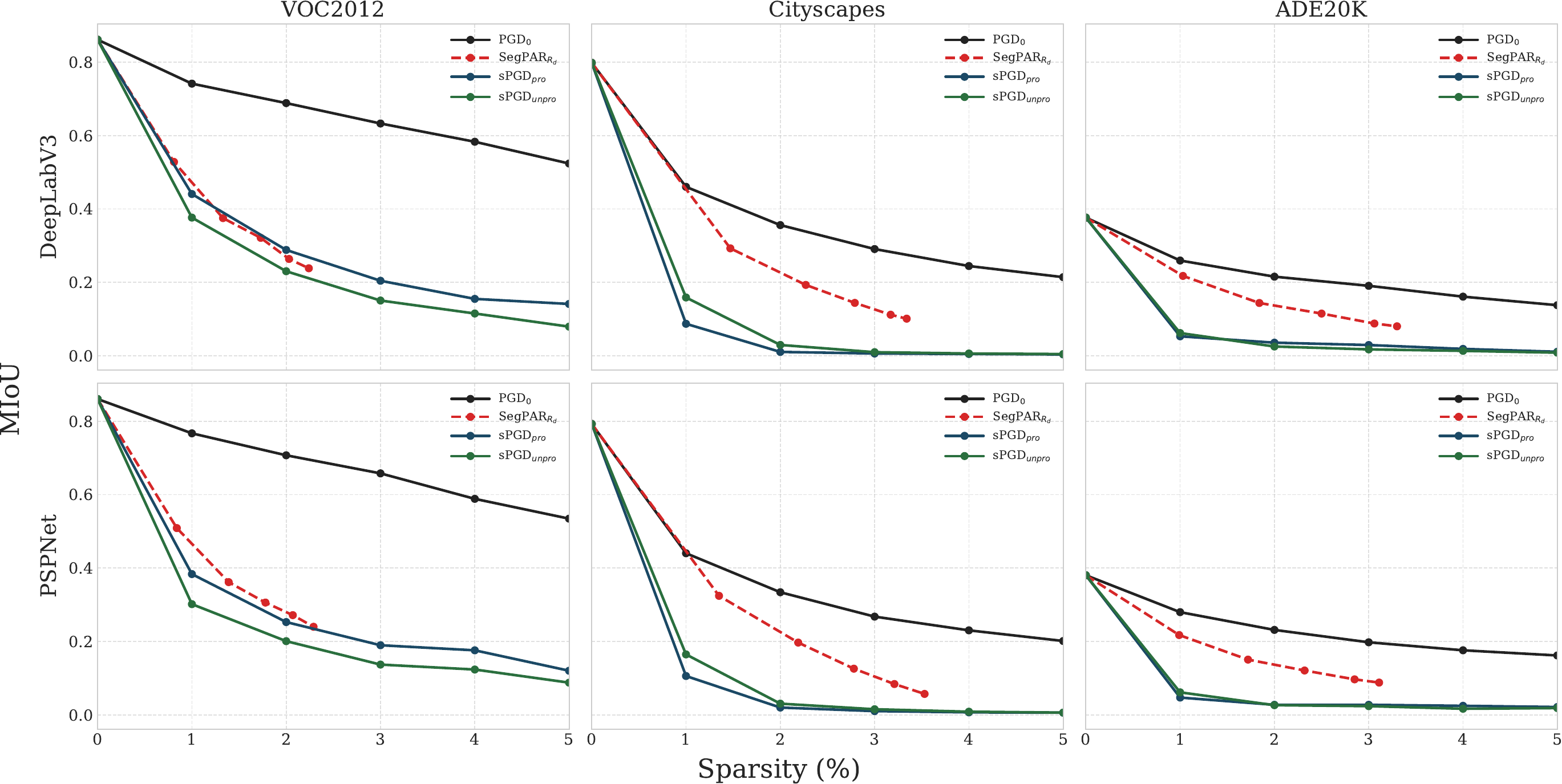}
  \caption{\textbf{Comparison of MIoU vs. Sparsity with White-box sparse Attacks.} The results demonstrate that our attack $\text{SegPAR}_{R_d}$  exhibits competitive performance relative to white-box sparse methods.}
  \label{fig:white-box comparison}
\end{figure}

In this section, we compare our attack, $\text{SegPAR}_{R_d}$, with white-box sparse baselines, $\text{PGD}_0$ \cite{PGD0}, $\text{sPGD}_{\text{pro}}$, and $\text{sPGD}_{\text{unpro}}$ \cite{sPGD}, all optimized with cross-entropy. For a standardized comparison, we align the evaluation budget by equating one black-box forward query to one white-box gradient update (which inherently entails both forward and backward passes). Under this metric, we enforce a maximum budget of 1,000 steps per image. Specifically, $\text{SegPAR}_{R_d}$ utilizes up to 1,000 forward queries, executed over 100 steps (where each mark represents a 20-step interval). Conversely, the white-box attacks allocate 200 gradient updates per sparsity level, totaling 1,000 updates across the 5 evaluated levels (\ie., $20$ restarts $\times$ $10$ iterations for $\text{PGD}_0$, and $200$ iterations for $\text{sPGD}$). As shown in \cref{fig:white-box comparison}, despite lacking access to internal gradients, $\text{SegPAR}_{R_d}$ remains strongly competitive across both segmentation models---often outperforming $\text{PGD}_0$ and approaching $\text{sPGD}$ under strict sparsity constraints. These results underscore the effectiveness of $\text{SegPAR}_{R_d}$ and suggest that stronger, segmentation-tailored white-box sparse attack baselines are still needed.

\section{Conclusion}
We proposed \textbf{SegPAR}, a decision-based black-box sparse attack for semantic segmentation.
By combining \textbf{class-centric exploration} with a \textbf{discrepancy reward} ($R_d$), SegPAR mitigates perturbation reversion and improves query efficiency in dense prediction, where the search space is vast.
Under strict query and sparsity budgets, $\text{SegPAR}_{R_d}$ achieves substantially larger MIoU degradation with fewer perturbed pixels than existing black-box baselines, while rapidly degrading the MIoU of safety-critical classes relevant to autonomous driving.
Even without access to internal model information, $\text{SegPAR}_{R_d}$ outperforms $\text{PGD}_0$ and narrows the gap to $\text{sPGD}$.
Furthermore, we observe that adversarial training can be less effective in lower-resolution regimes.
Overall, our findings highlight the persistent vulnerability of dense prediction models under realistic decision-only sparse threats.

\section*{Acknowledgements}
This work was supported by the Advanced GPU Utilization Support Program funded by the Ministry of Science and ICT (MSIT), Korea (No. 02-26-01-0134); by the National Research Foundation of Korea (NRF) grant funded by the Korean government (MSIT) (No. RS-2026-25498016); and by the BK21 FOUR Program through the NRF grant funded by the Korean government in 2025 (Grant No. 2120251815513), Institute for Sustainable VLEO Space Services Development.

%
%
\bibliographystyle{splncs04}
\bibliography{main}
\end{document}